\documentclass[10pt,twocolumn,letterpaper]{article}

\usepackage{cvpr} 
\usepackage{times}
\usepackage{epsfig}
\usepackage{amsmath}
\usepackage{amssymb}
\usepackage{float}
\usepackage{xcolor, colortbl}
\usepackage{multicol}
\usepackage{multirow}
\usepackage{graphicx}
\usepackage{url}
\usepackage{hyperref}
\hypersetup{
    colorlinks=true,
    linkcolor=blue,
    filecolor=magenta,      
    urlcolor=cyan,
    pdftitle={Overleaf Example},
    pdfpagemode=FullScreen,
    }

\begin{document}


\title{SynID: Passport Synthetic Dataset for Presentation Attack Detection}

\author{Juan E. Tapia, Fabian Stockhardt, Lázaro Janier González-Soler, Christoph Busch\\
da/sec-Biometrics and Internet Security Research Group, \\Hochschule Darmstadt (h-da), Germany. \\
{\tt\small juan.tapia-farias@h-da.de}
}

\maketitle
\thispagestyle{empty}

\begin{abstract}
The demand for Presentation Attack Detection (PAD) to identify fraudulent ID documents in remote verification systems has significantly risen in recent years. This increase is driven by several factors, including the rise of remote work, online purchasing, migration, and advancements in synthetic images. Additionally, we have noticed a surge in the number of attacks aimed at the enrolment process. Training a PAD to detect fake ID documents is very challenging because of the limited number of ID documents available due to privacy concerns. This work proposes a new passport dataset generated from a hybrid method that combines synthetic data and open-access information using the ICAO requirement to obtain realistic training and testing images.
\end{abstract}

\section{Introduction}
The need to detect fake ID documents has risen in the last years due to the increase in remote verification systems related to onboarding processes, pre-enrolment, online purchases, and other uses of smartphones \cite{Chen-tifs}. These processes are typically unsupervised, and the attacker can perform as many attacks as possible or as the system allows.

A remote verification system consists of two branches. The first branch determines whether the subject in front of the capture device is "live," meaning they are physically present in front of the smartphone. This captured image is a selfie taken under "wild" conditions. 

The second branch (focus of this work) evaluates the authenticity of each subject's ID card/passport. It checks to ensure that the ID card/passport is the original one issued by the official service and confirms that it has not been altered, modified or has been used in a replay attack in some way. A Presentation Attack Detection (PAD) system based on ID Documents is employed to assess the document's authenticity. 

During the last year, several initiatives and datasets have been proposed in the literature based mainly on a collection of ID cards and passports. However, existing datasets often lack the visual quality and structural accuracy necessary to simulate realistic identity forgeries in compliance with standard requirements. For instance, face images in ID cards/passports present the subject with a hat, sunglasses, rotation, and other inappropriate conditions for a genuine document \cite{KID34K}. The low resolution and quantity of the sample are also relevant factors, as is the case with the Prado dataset \footnote{\url{https://nidc.dk/en/Document-Database/ID-databases}}. The previously mentioned datasets' low quality makes them more suitable for testing a system instead of improving the performance, and they complement the training process of the PAD system.

Moreover, other limitations have been identified, such as the number of subjects and images. Many datasets present a video containing only a few subjects, but thousands of attack images are generated from the same subject. Because of the low number of subjects, the results are over-fitted and do not reach generalisation capabilities.

The bona fide images also present a challenge because, based on privacy limitations, plastic PVC card printed images are often used as bona fide, and attacks are generated from these bona fide images. However, the image components on the documents are not aligned with the enrolment rules and present unrealistic scenarios.

It is essential to highlight that the PAD system presents an asymmetrical relationship because the attacker only needs one well-explained fake ID document, compared to the defenders developing the detection algorithm, who need hundreds or thousands of bona fide images and attack samples to train a PAD system for a new threat.

Motivated by these previous limitations, we provide new passport datasets aligned with the ICAO requirements that will help to improve the PAD ID cards system's performance without compromising data protection.
This dataset was built in a hybrid mode, which means that synthetic images and online information are combined in an empty ID template.

The main contributions of this work are:
\begin{itemize}
    \item This is one of the First Passport datasets generated in a hybrid mode.
    \item The passport images (face, text and MRZ) are generated in compliance with ICAO, making the passport more realistic.
    \item This dataset is fully available for reproducibility upon request.
\end{itemize}

The rest of the paper is organised as follows: Section \ref{sec:related} reviews the related works. Section \ref{sec:method} explains the method, metrics \ref{sec:metric}. Section \ref{sec:expeandresults} explains the experiments and results obtained, and Section \ref{sec:conclusions} draws the conclusions of this work.

\section{Related Work}
\label{sec:related}

In the literature, only a few open-access datasets are available, such as MIDV500 \cite{MIDV500}, MIDV2020 \cite{MIDV2020}, DLC2021 \cite{MIDVDLC2021}, MIDV-Holo \cite{MIDVHolo}, Prado and IDNet \cite{IDNet}. Most of them are created from a low number of subjects with many images and different attacks and countries. However, stronger and more reliable results have been reported by private datasets with a high number of subjects, attacks and images per subject \cite{GONZALEZ2025111352}.

MIDVHolo dataset \cite{MIDVHolo} includes clips featuring documents, such as passports and ID cards, from the fictional country "Utopia." Only artificial documents were used for ethical reasons, which is why "Utopia" serves as a common name for this imaginary country. There are a total of 700 clips in the dataset. It was created to promote research on the topic of automatic hologram detection in video streams.

Benalcazar et al.\cite{benalcazar} proposed synthetic datasets for Chilean ID cards based on generative Adversarial networks and artificial textures. The results are very promising, showing that there is still room for improvement to generate high-quality images with more explicit text. The "Chocolate country" and "Tacoland" were created to demonstrate this technique.
\begin{figure*}[]
\centering
\includegraphics[scale=0.4]{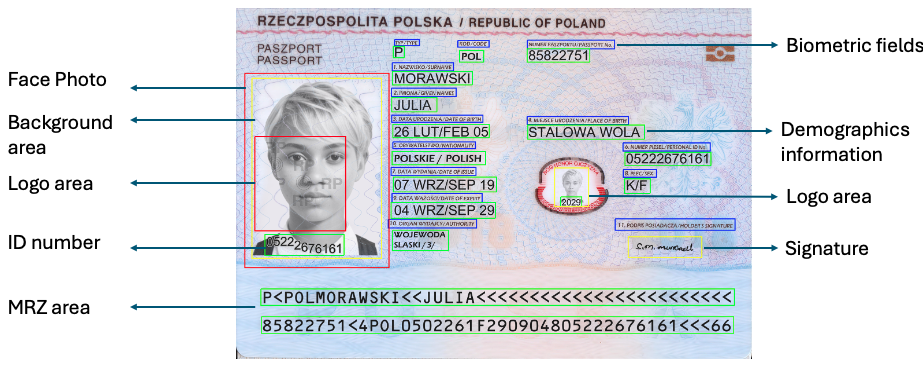}
\caption{Detailed images with all the changes highlighted in colours.}\label{fig:detail-pass}
\end{figure*}

In the KID34K \cite{KID34K} is a dataset created for online identity card fraud Detection. This dataset has ID cards with bona fide and print and screen attacks. No composite images are available. The dataset consists of a total of 34,662 images of 82 ID cards. For 46 subjects who did not exist, they produced 37 registration cards and 45 driver’s license cards. Of the 46 subjects, 36 have both types of ID cards. The dataset includes 13,746 genuine, 13,729 screens, and 7,187 print images. However, for the two real datasets, they collected only nine ID cards from nine volunteers and split the ID cards into two sets. The images are not compliant with ICAO requirements.

The IDNet \cite{IDNet} is a dataset proposed by the Homeland Security Department and includes Driver Licenses (DL), Passports and ID cards from several countries; also, the images are not compliant with ICAO requirements.  

Tapia et al. \cite{Tapia-IJCB2024} reported the results of the last IJCB 2024 competition on PAD on ID cards, highlighting the lack of images to train robust PAD systems and the lower capability to infer different countries, with a low performance to make this system feasible for real applications. 

Very recently, Gonzalez et al. \cite{GONZALEZ2025111352} showed good results that PAD systems can achieve when using real samples and a large number of private datasets, making this challenge more feasible. However, the private set is a limitation for improvement.

\section{Proposed Method}
\label{sec:method}

This section describes our pipeline for generating fake passports based on a hybrid method that includes open-access data and synthetic face images. This allows us to obtain visually convincing passport images. These passport images are ICAO-compliant and validated with a face image quality assessment software. 

The whole process is structured into five core components: (1) template normalization from layered Photoshop files, (2) generation of structured subject metadata, (3) biometric image selection and filtering, (4) multimodal compositing of image layers, and (5) reconstruction of complex visual overlays such as logos and patterns. While the implementation described here targets Polish passports, the pipeline generalises to other national formats with some adjustments in the JSON configuration files. Figure \ref{fig:detail-pass} shows all the areas of interest in one passport.

\subsection{ID-Template Recovery and Normalisation}

Passport templates used in this work were sourced from layered Photoshop (PSD) files that are publicly available via online websites or generated internally from an original passport\footnote{\url{https://www.pinterest.com/ideas/passport-documents/901672028621/}}. These ID templates, while widely circulated in digital forums and marketplaces, exhibit significant structural inconsistency across samples. In some cases, individual elements—such as biometric face regions, text labels, MRZ, and background patterns—are organised into distinct, editable layers. In other cases, content is flattened, mislabelled, or visually entangled across compositional artefacts. 

To enable scalable and automated generation, we developed a structured normalisation process to recover layer semantics and classify PSD components into consistent, interpretable categories.

We define four canonical layer classes, each specified explicitly in a country-specific configuration file provided as part of the framework: 
\begin{enumerate} 
    \item \textbf{Static Description Text}: fixed field labels such as “Name,” “Date of Birth,” and “Issuing Authority.” These are mostly invariant across samples. Some templates required colour correction or typographic adjustments to match reference passports. 
    \item \textbf{Subject Text Fields}: per-individual text content (e.g., names, ID numbers, birth dates), which is dynamically injected at generation time. These layers are fully overwritten during synthesis. 
    \item \textbf{Biometric area}: facial portraits, 
    and handwritten signatures. These elements are positioned and masked according to the document specification. 
    \item \textbf{Logos and Visual Patterns}: complex overlays, including watermark-like elements, security microtext, and background textures. These exhibit the highest variance and often require manual reconstruction. 
\end{enumerate} 

A configuration file (JSON) containing all the necessary information collected during this process. The configuration file serves as documentation and control logic for each document ID-template, allowing users to inspect, modify, or extend the layer mappings as needed for new document types.

Empty templates are derived by stripping subject-specific content while retaining spatial layout guides. Manual inspection and correction were necessary in all cases to ensure alignment fidelity with the authentic passport mock-up.

To guide this correction process, we compiled a collection of reference images from online sources, including forums, scan archives, and publicly shared samples of real passports. These references were visually inspected and cross-compared to identify common spatial layouts, typographic norms, and design motifs. While the legal provenance of these reference images could not be verified, our aim was to emulate the conditions under which a forger might operate—without privileged access to official design specifications or security documents. In practice, this reference set served as the ground truth against which we refined layer positions, fixed typographic errors, and reconstructed logo placements. 
This alignment was the most labour-intensive component of the pipeline, particularly as early-stage ID Templates suffered from malformed layout data.

\subsection{Subject Metadata Generation}

Subject-specific textual content was generated via a combination of curated lists and rule-based simulations. Public sources (e.g., Wikipedia) were used to construct culturally appropriate dictionaries for names (male/female first and last), cities, issuing authorities, and document numbers. Based on these lists, we instantiated 1050 synthetic individuals.

Machine-readable zone (MRZ) codes were generated using an open-source library \footnote{\url{https://pypi.org/project/mrz/}}, configured to match the ICAO 9303 specification. Character positioning, kerning offsets (distance between letters), and font metrics were modelled manually based on inspection, particularly for the non-machine-readable zones, which use country-specific and often non-standard fonts \footnote{\url{https://github.com/Arg0s1080/mrz}}.

All subject data and layout parameters were specified via country-specific configuration files. These files define not only the location of text fields and biometric images, but also document-level rendering parameters such as font families, kerning coefficients, colour values, transparency levels, and character alignment rules. 

\subsection{Biometric Data Selection and Filtering}

Facial portraits were sourced from the ONOT synthetic dataset \cite{onot}, which provides multiple synthetic face images per subject. 
However, the uncontrolled generation process yields images with variable pose, resolution, and artefacts. To approximate ICAO compliance, we developed a custom filtering pipeline based on the following criteria, which were validated by Open Face Image Quality (OFIQ) \cite{Merkle-OFIQ-Report-240930}:
\begin{itemize}
    \item Minimum face bounding box resolution
    \item Minimum relative eye-to-eye pixel distance
    \item Minimum relative margin from eye centre to image edge
    \item Sharpness according to OFIQ
\end{itemize}

No manual selection was performed; instead, the top-3 ranked images per subject were retained. 
Face detection and mesh estimation were performed using MediaPipe, while final segmentation was conducted in Adobe Photoshop due to its superior handling of edge blending and texture preservation in our tests. Face cropping was done according to the ICAO standard 9303. Mask and face mesh were saved and used in the later generation of the fake passport \footnote{\url{https://www.icao.int/publications/pages/publication.aspx?docnum=9303}}. 

\begin{figure}[H]
\centering
\includegraphics[scale=0.18]{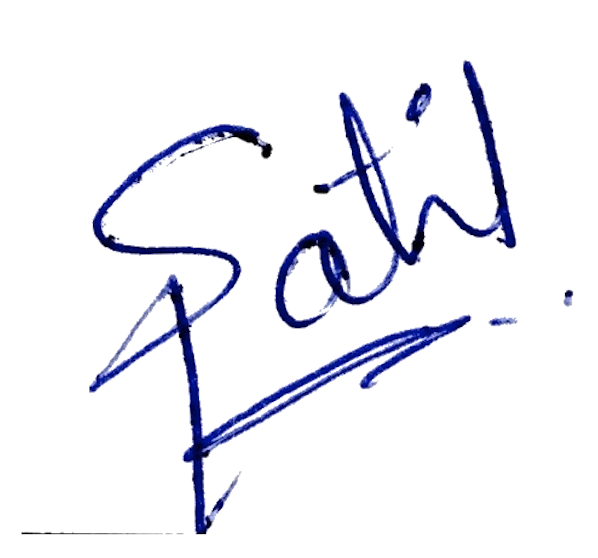}
\includegraphics[scale=0.18]{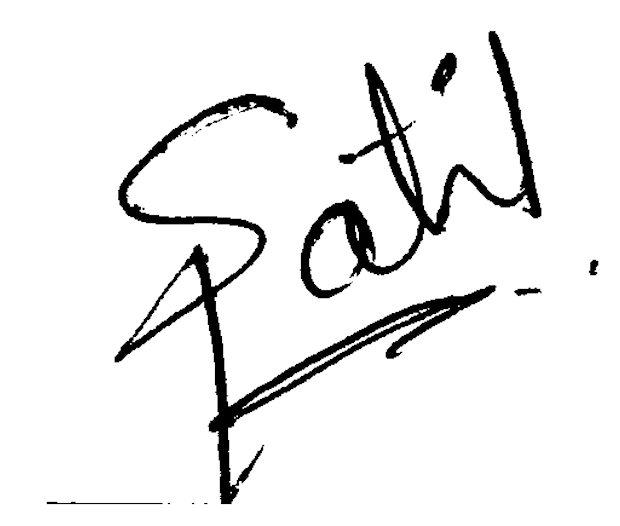}
\includegraphics[scale=0.18]{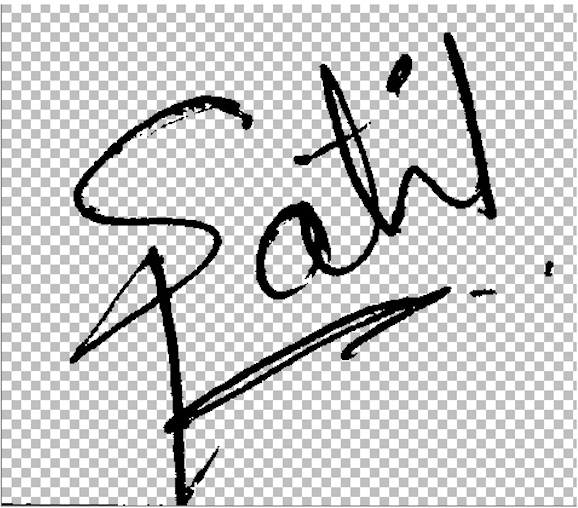}
\includegraphics[scale=0.3]{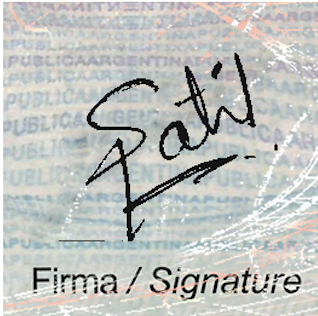}
\caption{Example of signature extraction process. Left to Right: Base image from the dataset. Binarise image. Remove background, add to passport.}
\label{fig:signature}
\end{figure}

Signatures were drawn from the OHSDA database \cite{Signature}. All biometric assets were preprocessed to remove backgrounds, with binary masks stored for compositing. Examples of the process are depicted in Figure \ref{fig:signature}.
Fingerprints were generated (for Argentinian or Colombian passports) using SFinGe and randomly assigned per subject. 

\subsection{Layer Compositing Pipeline}

Each synthetic passport was generated by compositing subject-specific data onto the cleaned ID-template. The placement and ordering of visual elements are controlled via layer masks and the rendering sequence defined in the configuration file. 

A range of post-processing operations is supported by the framework, of which the following were applied in the Polish passport setting:
\begin{itemize}
    \item \textbf{Edge blurring}: applied along mask boundaries to prevent hard seams.
    \item \textbf{Opacity tuning}: transparency levels are adjusted per-layer based on empirical inspection of authentic documents.
    \item \textbf{Character tuning}: character-level layout corrections are applied when the selected font does not fully match the visual appearance of the reference. This includes manual positioning, per-character kerning, and the application of predefined layout transformations.
\end{itemize}

Text fields are rendered using the visual characteristics extracted from the reference ID-template. These include font type, character spacing, rotation, and baseline curvature.

\subsection{Pattern and Logo Processing}

Visual elements such as holograms, emblems, and security patterns are either extracted from the template or reconstructed manually. When available, pattern layers were reused directly. Otherwise, we applied classical computer vision techniques (e.g., contour detection, colour thresholding) to extract patterns semi-automatically.

\begin{figure}[H]
\centering
\includegraphics[scale=0.05]{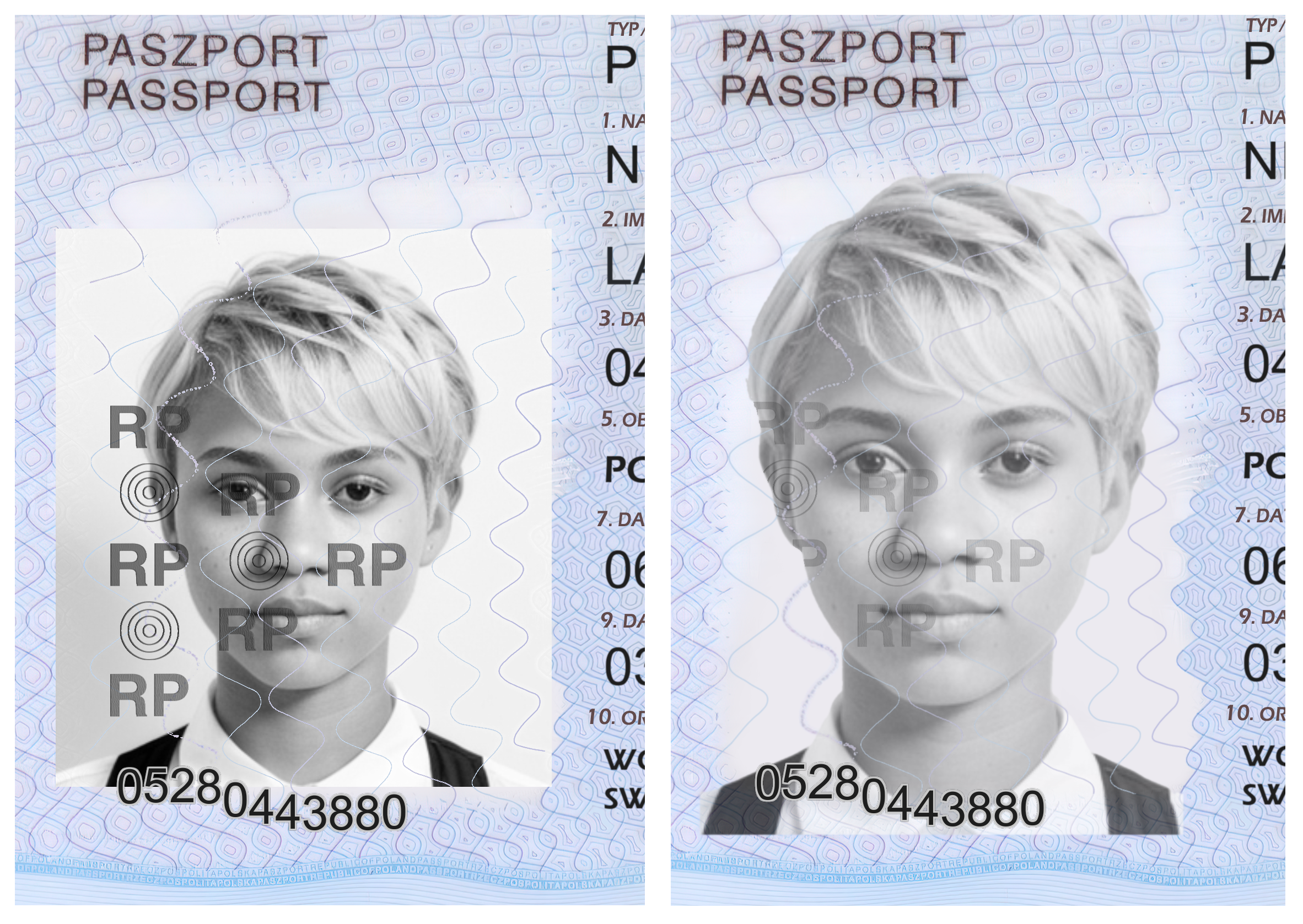}
\caption{Detailed logo and pattern information created in order to follow the pattern of the Polish passport.}\label{fig:detail-logo}
\end{figure}

In challenging cases—where the template did not provide a cleanly segmented logo layer, and all available reference examples showed the logo partially occluded by text, biometric data, or other overlays—segmentation was performed manually using GIMP. Colours were corrected via palette estimation, and placement masks were constructed to guide compositing logic. Figure \ref{fig:detail-logo} depicts an example applied to the logo adaptation of the face picture.

Three empty templates were used to generate fake Passports belonging to Spain, Portugal, and Poland. All the passports generated follow the original pattern presented by each country.

\begin{figure*}[]
\centering
\includegraphics[width=\textwidth]{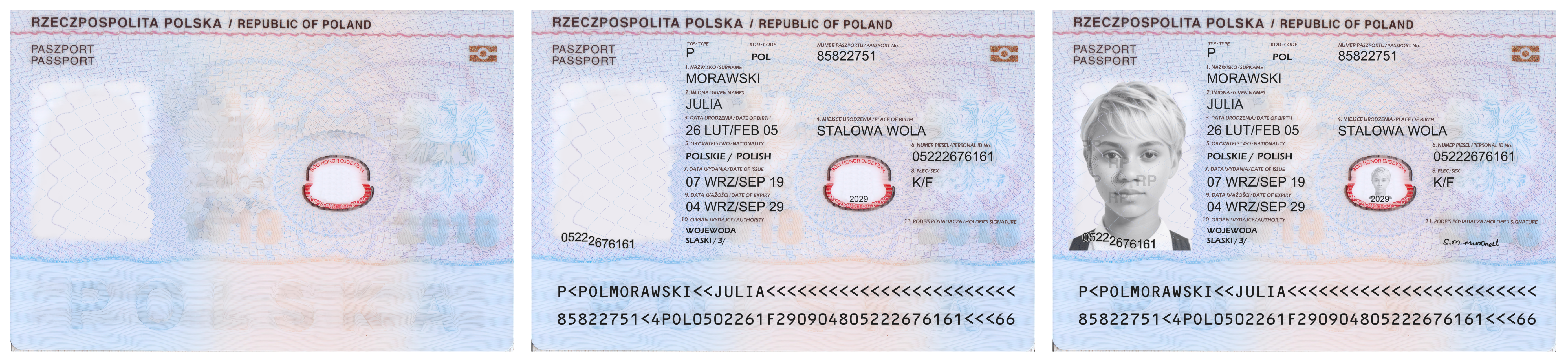}
\caption{Basic steps for fake passport generation. Left to right:  Empty template, intermediate state with demographic text, and final composition with biometric image.}
\label{fig:3-steps}
\end{figure*}

Figure \ref{fig:3-steps} three passport images that describe the
full process from an empty template up to the final version.

\begin{figure*}
    \centering
    \includegraphics[scale=0.13]{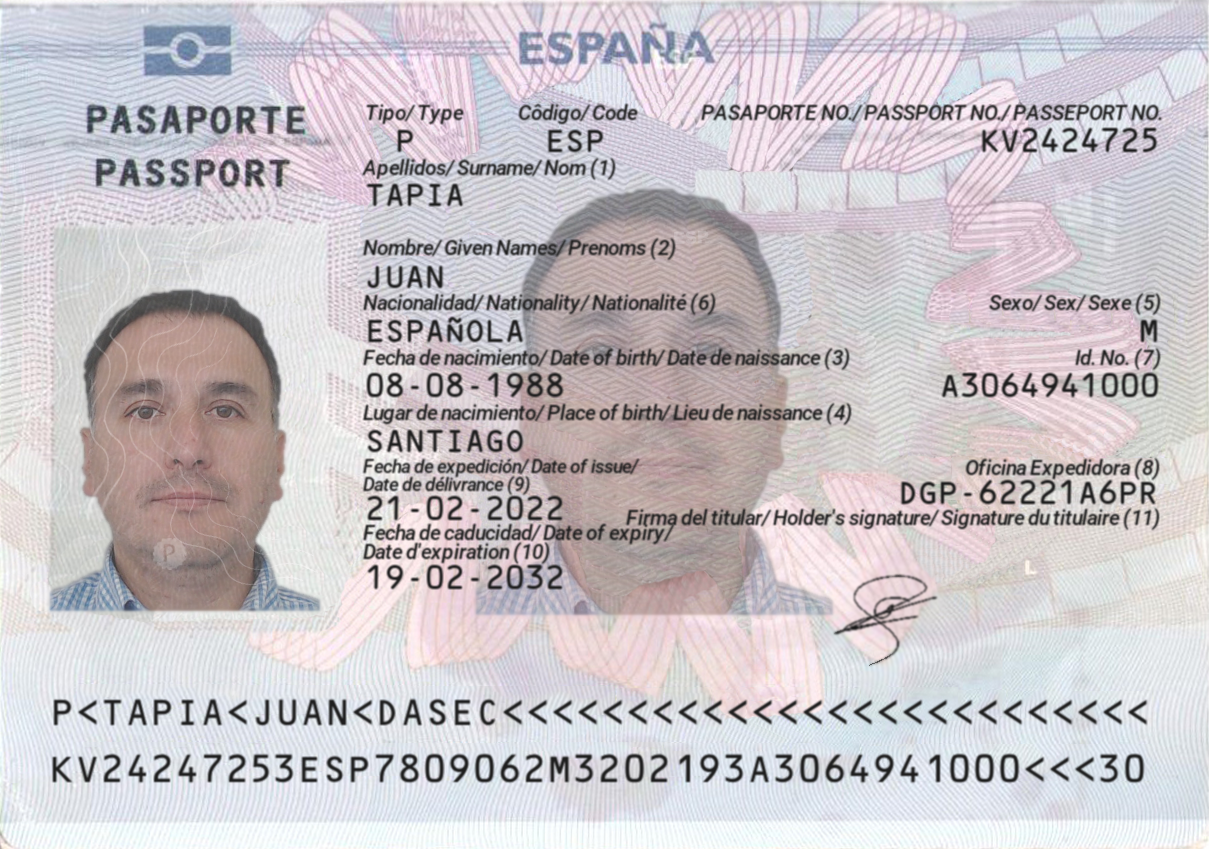}
    \includegraphics[scale=0.045]{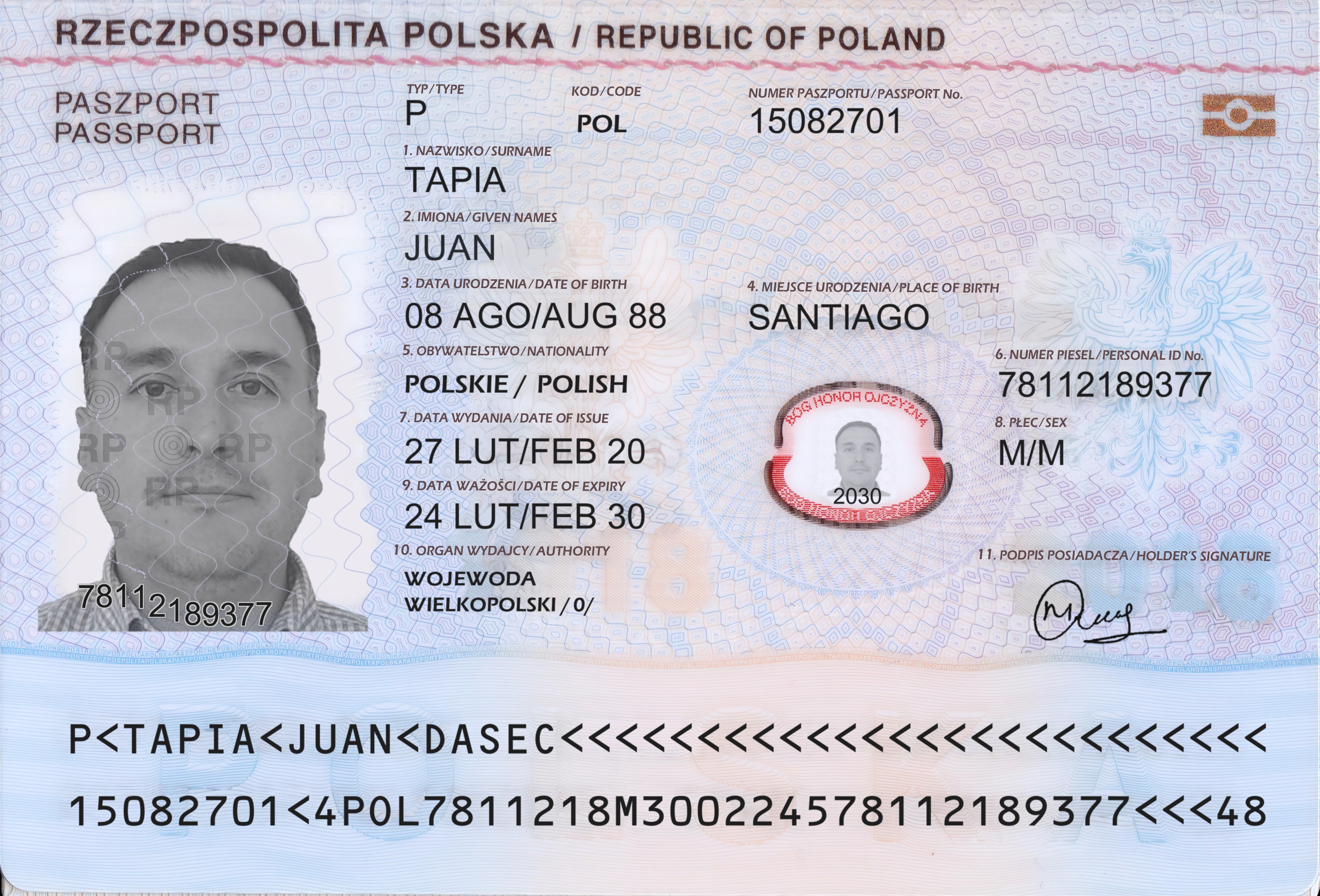}
    \includegraphics[scale=0.225]{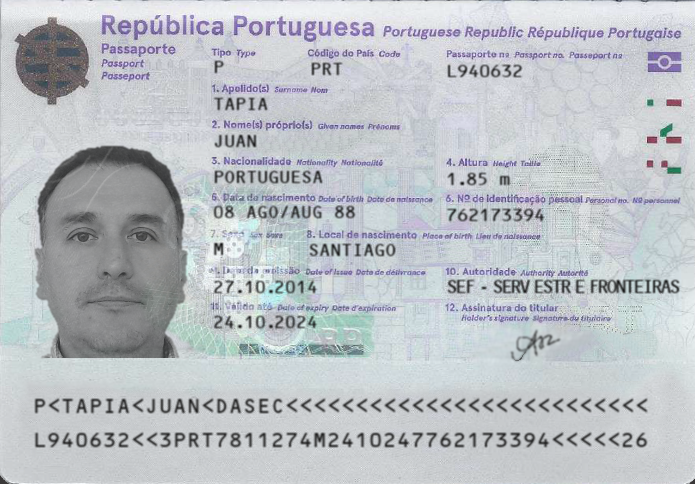}
    \caption{Example of three fake passport generations for the following countries. Left to right: Spain, Poland, Portugal.} 
    \label{fig:pass-example}
\end{figure*}

Figure \ref{fig:pass-example} presents an example of each passport generated for Spain, Poland and Portugal.

\section{Metrics}
\label{sec:metric}

The ISO/IEC 30107-3 standard\footnote{\url{https://www.iso.org/standard/67381.html}} presents methodologies for evaluating the performance of PAD algorithms for biometric systems. The APCER metric measures the proportion of attack presentations---for each different Presentation Attacks Instrument (PAI)---incorrectly classified as bona fide presentations. This metric is calculated for each PAI, where the worst-case scenario is considered. Equation~\ref{eq:apcer} details how to compute the APCER metric, in which the value of $N_{PAIS}$ corresponds to the number of attack presentation images, where $RES_{i}$ for the $i$th image is $1$ if the algorithm classifies it as an attack presentation, or $0$ if it is classified as a bona fide presentation (real image).

\begin{equation}\label{eq:apcer}
    {APCER_{PAIS}}=1 - (\frac{1}{N_{PAIS}})\sum_{i=1}^{N_{PAIS}}RES_{i}
\end{equation}

Additionally, the BPCER metric measures the proportion of bona fide presentations mistakenly classified as attack presentations or the ratio between false rejection and total bona fide attempts. The BPCER metric is formulated according to equation~\ref{eq:bpcer}, where $N_{BF}$ corresponds to the number of bona fide presentation images, and $RES_{i}$ takes identical values to those of the APCER metric.

\begin{equation}\label{eq:bpcer}
    BPCER=\frac{\sum_{i=1}^{N_{BF}}RES_{i}}{N_{BF}}
\end{equation}

These metrics effectively measure to what degree the algorithm confuses presentations of attack images with bona fide images and vice versa. The APCER and BPCER metrics depend on a decision threshold.

\begin{figure*}[]
\centering
\includegraphics[scale=0.27]{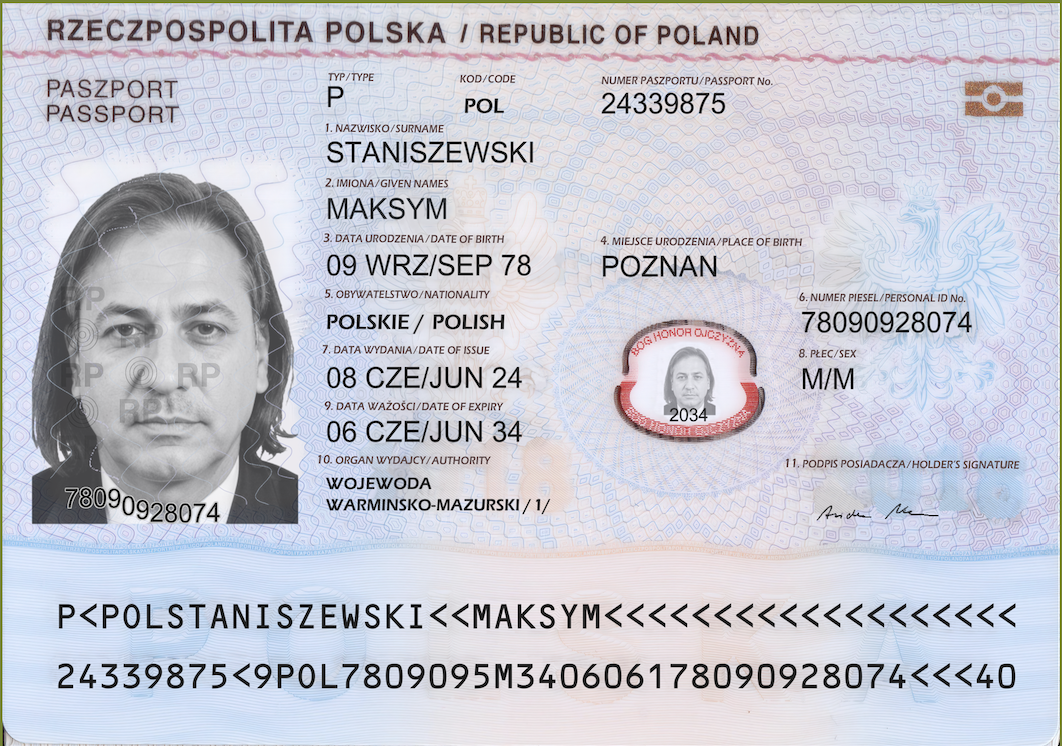}
\includegraphics[scale=0.20]{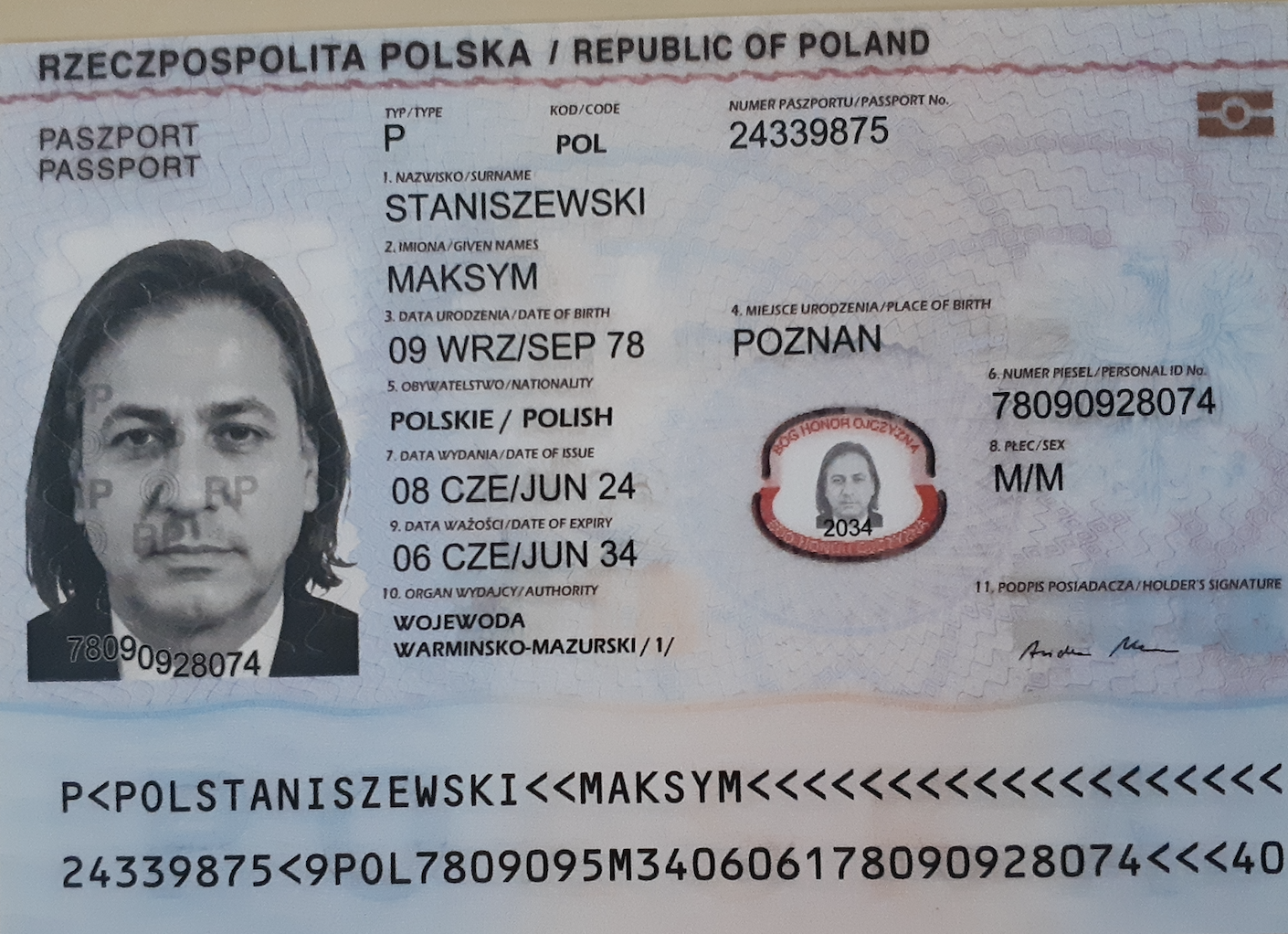}
\includegraphics[scale=0.265]{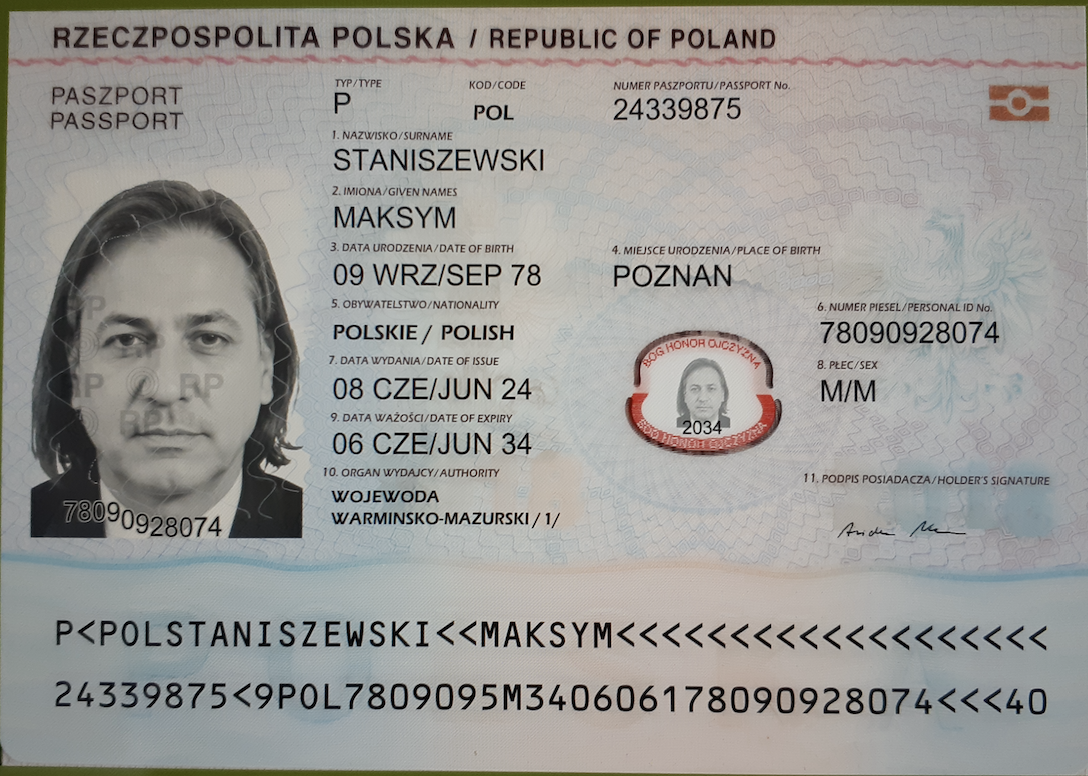}
\caption{Examples of bona fide, printed and screen images generated for PAD system.}\label{fig:attacks}
\end{figure*}

\section{Experiment and Results}
\label{sec:expeandresults}

Two experiments, a binary classifier and a leave-one-out protocol, were proposed for evaluating our passport synthetic datasets. 

In order to train a PAD classifier, 1000 images were automatically generated for each country. In total, we have 3000 images.

For the three countries' passport images generated, the print and screen images were manually created as attacks one by one. For printed attack, four images per sheet were placed on a glossy 180-gram paper to keep the official size at 600 dpi and simulate the original passport materials. For screen attack, the images generated were displayed on a screen and captured using a smartphone. An example of bona fide, print and screen is shown in Figure \ref{fig:attacks}.

In total, the dataset has 9.000 images divided into 3000 bona fide (Passport generated), 3000 manually printed attacks captured with two different smartphones, and 3000 manually screened attacks captured from Dell laptop screens.

The full dataset for intra-dataset was divided into 60\% for training, 20\% for validation and 20\% for test.

For the leave-one-out protocol, the 3000 images belonging to Poland (bona fide, print and screen) were used as a test set, and Spain and Portugal were used as training and validation subsets. The print and screen attacks were evaluated separately. 

For the intra-dataset evaluation, the input size was adapted according to each model for deep learning and visual transformer models.
 
In the case of SwinT models, the most detailed configuration is described. The patch size division that makes a stronger classifier with the follow configuration: image-size = $224$, patch-size = $4$, num-channels = $3$, embed-dim = $96$, depths = [2, 2, 6, 2], num-heads = [3, 6, 12, 24], window-size = $7$, mlp-ratio = $4.0$, drop-path-rate = $0.1$, hidden-act = 'gelu', initializer-range = $0.02$, layer-norm-eps = $1e-05$ encoder-stride = $3$.

The parameters for all SwinT are selected according to:
SwinT-B, SwinT-S, and SwinT-L model sizes, whose complexities are increased from baseline SwinT-B by $\times0.5$,$\times1$, and $\times2$. Window size $M$ is set to $7$ by default, and the query dimension of each head(as multi-head attention is used) is $32$ for all these models.


\begin{figure*}[]
\centering
\includegraphics[scale=0.28]{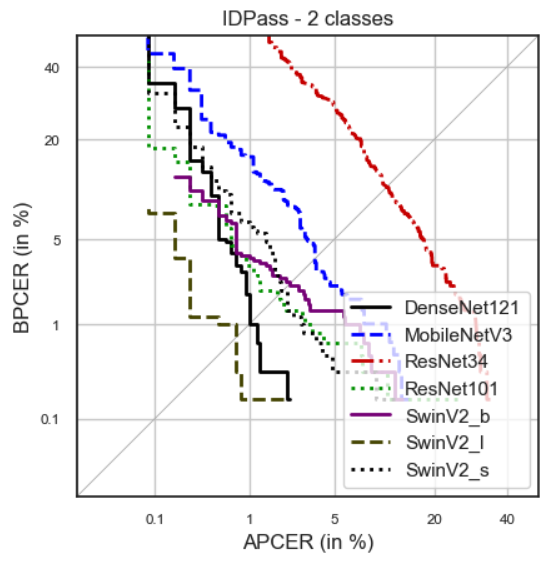}
\includegraphics[scale=0.28]{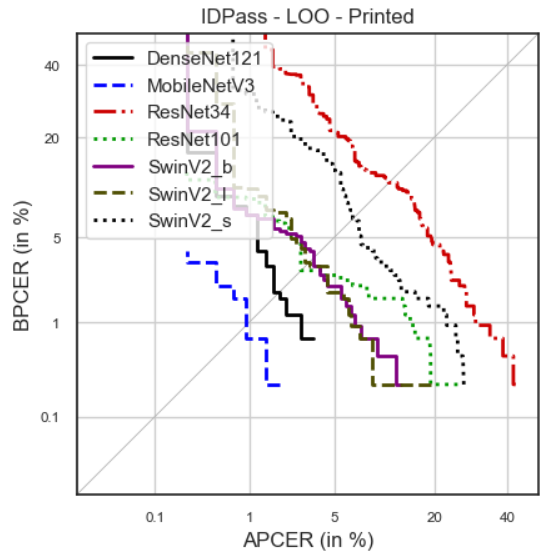}
\includegraphics[scale=0.28]{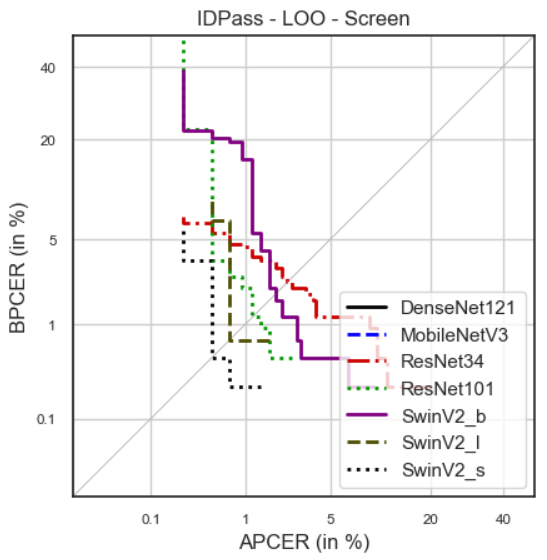}
\caption{DET curve shows the results for a deep learning and SwinT classifiers considering the three countries. Left: in intra-dataset evaluation. Middle: LOO print attack as a test. Right: LOO print screen attacks as test.}\label{fig:det-DL}
\end{figure*}

\begin{table*}[]
\scriptsize
\centering
\caption{PAD summary results. Intra and LOO scenarios considering Portugal and Spain as the training set. The Polish passport images are used as a test set.}
\label{tab:loo-eval-pol}
\resizebox{\textwidth}{!}{%
\begin{tabular}{|ccccclccccccc|}
\hline
\multicolumn{1}{|c|}{\multirow{2}{*}{Models}} &
  \multicolumn{4}{c||}{Intra} &
  \multicolumn{4}{c|}{LOO - Printed} &
  \multicolumn{4}{c|}{LOO - Screen} \\ \cline{2-13} 
\multicolumn{1}{|c|}{} &
  \multicolumn{1}{c|}{\begin{tabular}[c]{@{}c@{}}EER\\ (\%)\end{tabular}} &
  \multicolumn{1}{c|}{\begin{tabular}[c]{@{}c@{}}BPCER10\\ (\%)\end{tabular}} &
  \multicolumn{1}{c|}{\begin{tabular}[c]{@{}c@{}}BPCER20\\ (\%)\end{tabular}} &
  \multicolumn{1}{c||}{\begin{tabular}[c]{@{}c@{}}BPCER100\\ (\%)\end{tabular}} &
  \multicolumn{1}{c|}{\begin{tabular}[c]{@{}c@{}}EER\\ (\%)\end{tabular}} &
  \multicolumn{1}{c|}{\begin{tabular}[c]{@{}c@{}}BPCER10\\ (\%)\end{tabular}} &
  \multicolumn{1}{c|}{\begin{tabular}[c]{@{}c@{}}BPCER20\\ (\%)\end{tabular}} &
  \multicolumn{1}{c|}{\begin{tabular}[c]{@{}c@{}}BPCER100\\ (\%)\end{tabular}} &
  \multicolumn{1}{c|}{\begin{tabular}[c]{@{}c@{}}EER\\ (\%)\end{tabular}} &
  \multicolumn{1}{c|}{\begin{tabular}[c]{@{}c@{}}BPCER10\\ (\%)\end{tabular}} &
  \multicolumn{1}{c|}{\begin{tabular}[c]{@{}c@{}}BPCER20\\ (\%)\end{tabular}} &
  \begin{tabular}[c]{@{}c@{}}BPCER100\\ (\%)\end{tabular} \\ \hline
\multicolumn{13}{|c|}{Deep Learning} \\ \hline
\rowcolor{gray!30}\multicolumn{1}{|c|}{DenseNet121} &
  \multicolumn{1}{c|}{1.39} &
  \multicolumn{1}{c|}{0.3} &
  \multicolumn{1}{c|}{0.53} &
  \multicolumn{1}{c||}{2.39} &
  \multicolumn{1}{l|}{1.79} &
  \multicolumn{1}{c|}{0.0} &
  \multicolumn{1}{c|}{0.0} &
  \multicolumn{1}{c|}{7.20} &
  \multicolumn{1}{c|}{0.0} &
  \multicolumn{1}{c|}{0.0} &
  \multicolumn{1}{c|}{0.0} &
  0.0 \\ \hline
\multicolumn{1}{|c|}{MobilenetV3-l} &
  \multicolumn{1}{c|}{3.54} &
  \multicolumn{1}{c|}{1.03} &
  \multicolumn{1}{c|}{2.16} &
  \multicolumn{1}{c||}{16.33} &
  \multicolumn{1}{l|}{0.81} &
  \multicolumn{1}{c|}{0.0} &
  \multicolumn{1}{c|}{0.0} &
  \multicolumn{1}{c|}{0.69} &
  \multicolumn{1}{c|}{0.0} &
  \multicolumn{1}{c|}{0.0} &
  \multicolumn{1}{c|}{0.0} &
  0.0 \\ \hline
\multicolumn{1}{|c|}{ResNet34} &
  \multicolumn{1}{c|}{11.0} &
  \multicolumn{1}{c|}{13.33} &
  \multicolumn{1}{c|}{28.51} &
  \multicolumn{1}{c||}{60.03} &
  \multicolumn{1}{l|}{11.36} &
  \multicolumn{1}{c|}{12.55} &
  \multicolumn{1}{c|}{21.16} &
  \multicolumn{1}{c|}{53.72} &
  \multicolumn{1}{c|}{2.31} &
  \multicolumn{1}{c|}{0.93} &
  \multicolumn{1}{c|}{1.16} &
  4.41 \\ \hline
\multicolumn{1}{|c|}{ResNet101} &
  \multicolumn{1}{c|}{1.83} &
  \multicolumn{1}{c|}{0.33} &
  \multicolumn{1}{c|}{0.67} &
  \multicolumn{1}{c||}{3.16} &
  \multicolumn{1}{l|}{2.78} &
  \multicolumn{1}{c|}{1.62} &
  \multicolumn{1}{c|}{2.55} &
  \multicolumn{1}{c|}{9.06} &
  \multicolumn{1}{c|}{1.15} &
  \multicolumn{1}{c|}{0.0} &
  \multicolumn{1}{c|}{0.0} &
  2.09 \\ \hline
\multicolumn{13}{|c|}{Vision Transformers} \\ \hline
\multicolumn{1}{|c|}{Swin\_v2\_b} &
  \multicolumn{1}{c|}{2.20} &
  \multicolumn{1}{c|}{0.33} &
  \multicolumn{1}{c|}{1.44} &
  \multicolumn{1}{c||}{3.67} &
  \multicolumn{1}{l|}{3.36} &
  \multicolumn{1}{c|}{0.46} &
  \multicolumn{1}{c|}{2.09} &
  \multicolumn{1}{c|}{7.20} &
  \multicolumn{1}{c|}{1.73} &
  \multicolumn{1}{c|}{0.0} &
  \multicolumn{1}{c|}{0.46} &
  15.81 \\ \hline
\multicolumn{1}{|c|}{Swin\_v2\_s} &
  \multicolumn{1}{c|}{2.10} &
  \multicolumn{1}{c|}{0.17} &
  \multicolumn{1}{c|}{0.34} &
  \multicolumn{1}{c||}{6.17} &
  \multicolumn{1}{l|}{6.72} &
  \multicolumn{1}{c|}{3.25} &
  \multicolumn{1}{c|}{13.02} &
  \multicolumn{1}{c|}{31.16} &
  \multicolumn{1}{c|}{0.46} &
  \multicolumn{1}{c|}{0.0} &
  \multicolumn{1}{c|}{0.0} &
  0.23 \\ \hline
\rowcolor{gray!30}\multicolumn{1}{|c|}{Swin\_v2\_t} &
  \multicolumn{1}{c|}{0.70} &
  \multicolumn{1}{c|}{0.0} &
  \multicolumn{1}{c|}{0.0} &
  \multicolumn{1}{c||}{0.17} &
  \multicolumn{1}{l|}{3.13} &
  \multicolumn{1}{c|}{0.23} &
  \multicolumn{1}{c|}{1.86} &
  \multicolumn{1}{c|}{10.46} &
  \multicolumn{1}{c|}{0.69} &
  \multicolumn{1}{c|}{0.0} &
  \multicolumn{1}{c|}{0.0} &
  0.69 \\ \hline
\end{tabular}%
}
\end{table*}

\subsection{Experiment 1 - Binary classifier}
In order to show the generalization capabilities of deep learning models, we evaluated five classifiers trained from scratch based on ImageNet weights as a baseline on two classes bona fide and attacks. The two best methods are highlighted in a grey colour background. 

The networks selected were: DenseNet121\cite{DenseNet}, MobileNetV3-large \cite{mbv3}, ResNet34 and ResNet101 \cite{resnet}. Also, a Swin-Transformer \cite{swin} was used based on Vision Transformer capabilities that make a difference with traditional Deep learning models. The models evaluated are: SwinT-Base, SwinT-Small, and SwinT-large. 

All the models were trained using SGD and Adam optimisers. Different learning rates were explored using a grid search from $1e-3$ to $1e-6$. The $lr$ value of $1e-3$ and $50$ epochs achieved the best results.

Figure \ref{fig:det-DL} shows the DET curves for Intra-dataset (left) and leave-one-out protocol (middle and right). The three countries, Spain, Portugal, and Poland, were used in intra-datasets and divided into two classes, bona fide versus attacks. The best results are obtained by DenseNet121 for the DL models and SwinT-L for the Vision Transformer models. Conversely, the ResNet34 obtained the worst results in terms of EER.

\subsection{Experiment 2 - Leave-one-out classifier}
For the LOO protocol, we applied the following order for Spain, Portugal, and Poland passports, considering print and screen attacks.
For the training process, Spain and Portugal were used for training and validation, and Poland for testing. The attacks were evaluated separately to emulate unknown attacks.

For the LOO protocol, the evaluation was made considering bona fide versus print and screen attacks. Figure \ref{fig:det-DL}(middle) shows the results for the LOO for the Poland passport as a test set, considering a printed attack using it as an unknown attack. The model with the best results was MobileNetV3, and the worst was ResNet34.

Figure \ref{fig:det-DL}(right) shows the results for the LOO for screen attack using it as an unknown attack. The model with the best results was SwinV2-S and the worst ResNet34.

Table \ref{tab:loo-eval-pol} shows a summary of the results for both protocols and includes three different operational points, such as BPCER10, BPCER20 and BPCER100, considering traditional deep learning classifiers and Swin-T based on Vision-Transformers. As expected, the intra-dataset scenario obtained a low EER because bona fide and attacks were created using the same techniques.
For the attacks, the screen scenarios are identified as less difficult in comparison to the printed ones.

\subsection{Benchmark with State of the art}

In order to evaluate whether the traditional PAD-ID card system is vulnerable with respect to the generated images, all the passport images generated were used to benchmark two state-of-the-art systems.
The first model trained was based on a subset used in \cite{GONZALEZ2025111352}, which contains Chilean ID Cards for training, considering print and attack scenarios. The second model trained was based on the IDNet dataset for training \cite{IDNet}, which includes a passports subset with printed and screen scenarios, among others.

\begin{figure}[H]
\centering
\includegraphics[scale=0.21]{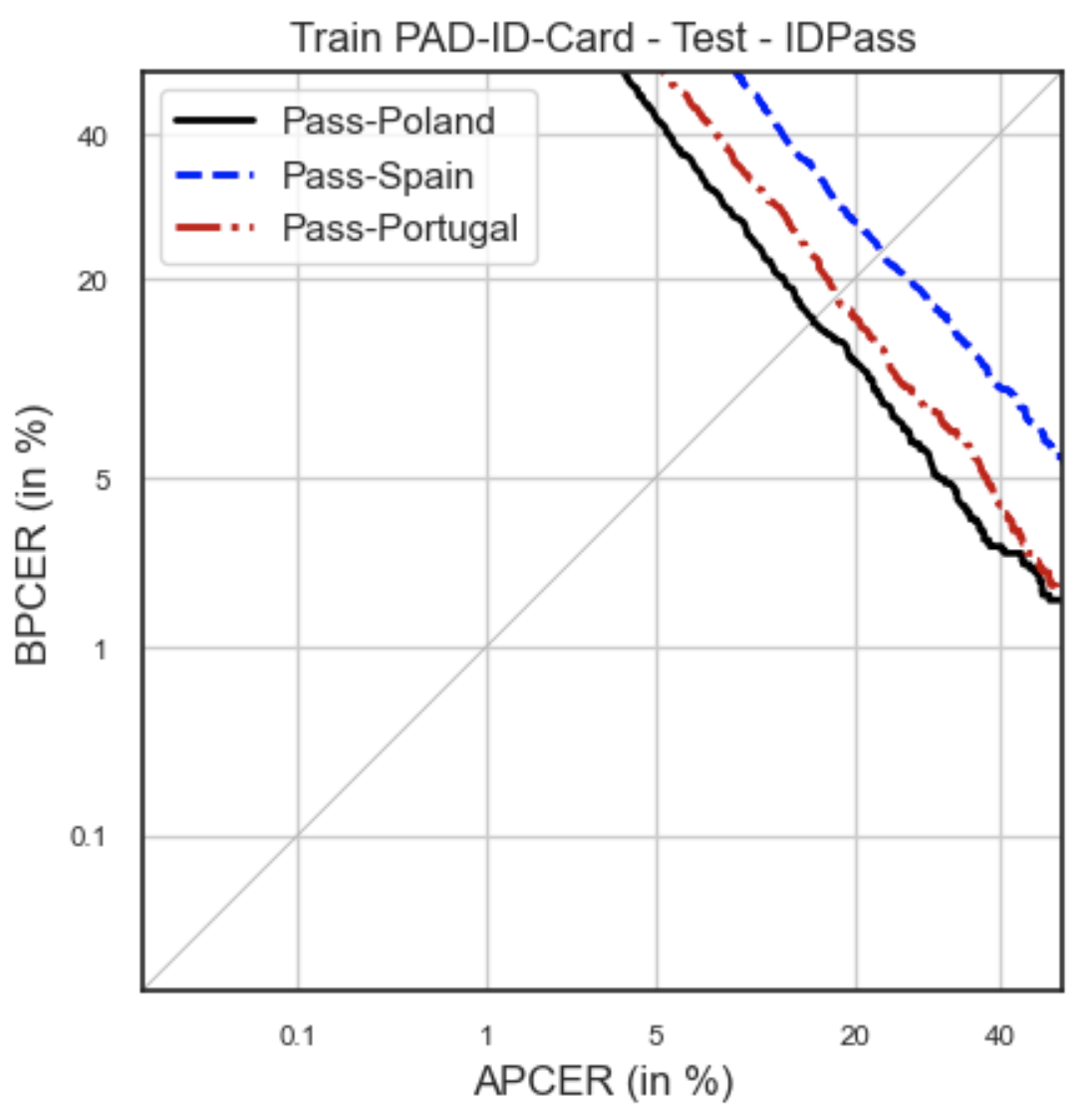}
\includegraphics[scale=0.21]{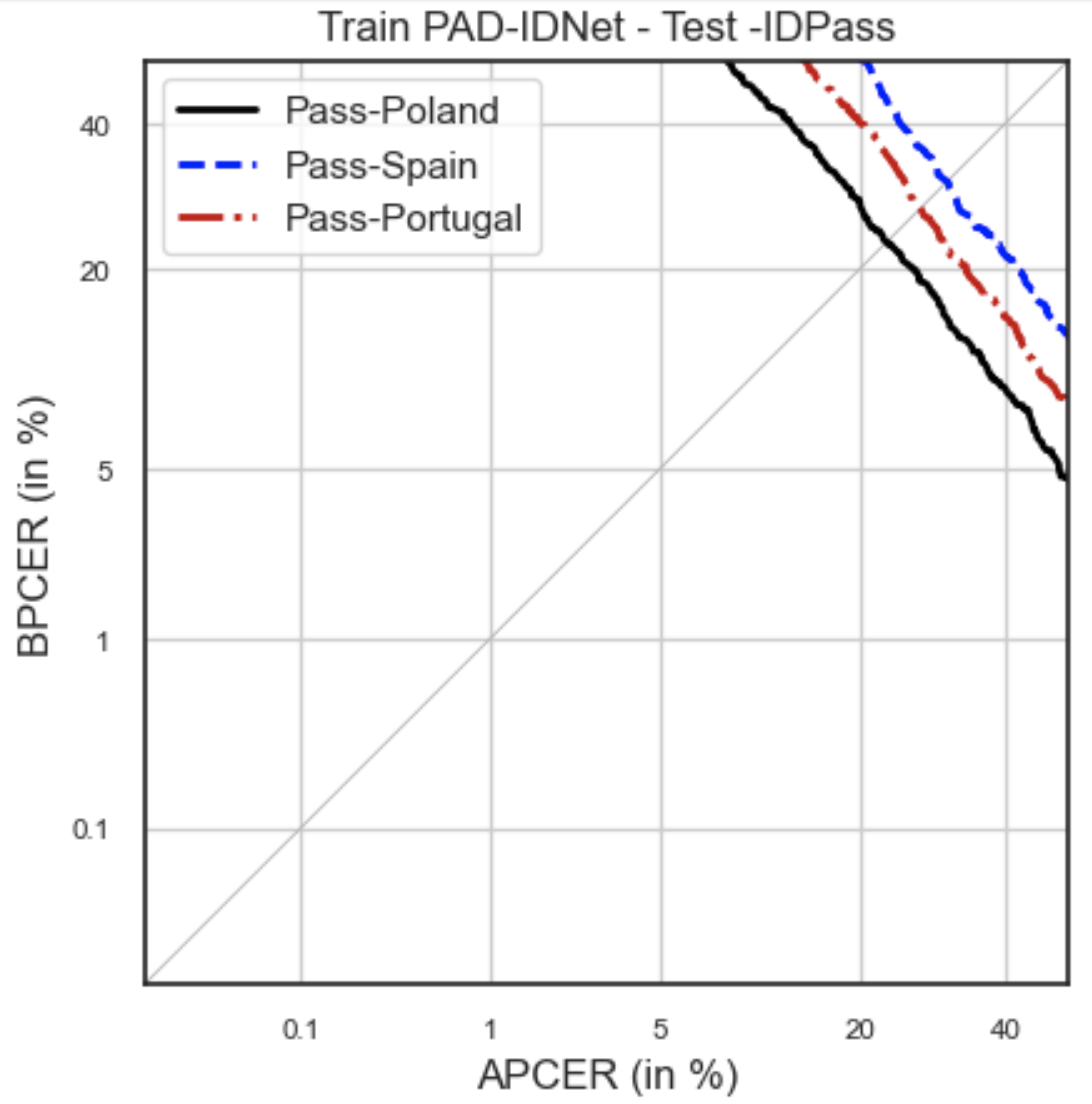}
\caption{DET curve shows the results for three countries generated when evaluated in a state-of-the-art classifier considering Chilean ID card and IDNet-Passport sets.}\label{fig:eval-IDcard-IDNet}
\end{figure}

Figure \ref{fig:eval-IDcard-IDNet}, shows the DET curve results for the three datasets generated by countries that include printed and screen attacks. As depicted, the passports generated are very challenging even for the two stronger classifiers aligned with our goal. For both evaluations, the Polish passport (black line) is identified as more complex than Spain (blue line) and Portugal (red line). The EER is over 17.00\% for PAD-ID and over 20\% IDNet.

It is essential to highlight that real systems evaluate ID Cards and Passport images in the same pipeline. The companies usually do not generate separate models for each one.

\section{Conclusion}
\label{sec:conclusions}
The method utilising SwinTransformer achieved the best results, highlighting the effectiveness of dividing images into small patches based on attention map layers. This approach allows the network to concentrate on specific areas, enhancing its ability to detect changes within each region. In contrast, while deep learning convolution techniques can capture texture changes at various resolutions, this capability is insufficient to enhance the PAD system.
As a future work, we will expand this research and dataset for other countries, including new attacks such as composite and plastic.

\section*{Acknowledgments}
This research work has been partially funded by the European Union (EU) under G.A. 101121280 (EINSTEIN) and CarMen (101168325), and UKRI Funding Service under IFS reference 10093453, and the German Federal Ministry of Education and Research and the Hessian Ministry of Higher Education, Research, Science and the Arts within their joint support of the National Research Center for Applied Cybersecurity ATHENE.
\newpage
{\small
\bibliographystyle{ieee}
\bibliography{main}
}

\end{document}